\documentclass[runningheads]{llncs}
\usepackage[T1]{fontenc}
\usepackage{graphicx}
\graphicspath{{figures/}}
\usepackage{color}
\usepackage{enumitem}
\usepackage{multirow}
\usepackage{multicol}
\usepackage{array}
\usepackage{subcaption,ragged2e}
\usepackage{framed, amsmath, amssymb}
\usepackage{booktabs}
\usepackage{float}
\newcolumntype{C}[1]{>{\centering\arraybackslash}p{#1}}
\newcolumntype{?}{!{\vrule width 1pt}}

\usepackage{adjustbox}

\begin{document}
\title{Language Models are Few-Shot Graders}
%
%



\author{Chenyan Zhao\inst{1}\orcidID{0000-0001-9520-2438} \and
Mariana Silva\inst{1}\orcidID{0000-0001-6087-7179} \and
Seth Poulsen\inst{2}\orcidID{0000-0001-6284-9972}}
\authorrunning{Zhao et al.}
%
\institute{University of Illinois Urbana-Champaign\and
Utah State University}
\maketitle              
\begin{abstract}

Providing evaluations to student work is a critical component of effective student learning, and automating its process can significantly reduce the workload on human graders. Automatic Short Answer Grading (ASAG) systems, enabled by advancements in Large Language Models (LLMs), offer a promising solution for assessing and providing instant feedback for open-ended student responses. In this paper, we present an ASAG pipeline leveraging state-of-the-art LLMs.  
Our new LLM-based ASAG pipeline achieves better performances than existing custom-built models on the same datasets. 
We also compare the grading performance of three OpenAI models: GPT-4, GPT-4o, and o1-preview. Our results demonstrate that GPT-4o achieves the best balance between accuracy and cost-effectiveness. On the other hand, o1-preview, despite higher accuracy, exhibits a larger variance in error that makes it less practical for classroom use. 
We investigate the effects of incorporating instructor-graded examples into prompts using no examples, random selection, and Retrieval-Augmented Generation (RAG)-based selection strategies. Our findings indicate that providing graded examples enhances grading accuracy, with RAG-based selection outperforming random selection. Additionally, integrating grading rubrics improves accuracy by offering a structured standard for evaluation. 

\end{abstract}

\section{Introduction and Background}
Providing grading and feedback has been shown to be highly effective in enhancing student learning outcomes~\cite{shute2008feedback}. However, manually grading assessments is one of the most time-consuming and tedious aspects of instruction, making it difficult to scale~\cite{bonthu2021automated,henkel2024can}. To alleviate this burden, educators often turn to closed-ended question formats, such as multiple-choice or numerical input questions, which can be automatically graded~\cite{pearson2005assessment}. While these formats enable efficient grading and feedback, they also present limitations. For example, students may develop lower-level skills and rely on test-taking strategies rather than deeply engaging with the material~\cite{ashish2024tradeoff}. Additionally, instructors face challenges in designing high-quality distractors and effective closed-ended questions~\cite{scott2006evaluating}.

In contrast, open-ended questions allow students to express their understanding in their own words, providing a more accurate measure of their knowledge~\cite{ashish2024tradeoff}. However, grading such responses at scale remains a significant challenge. Automated Short Answer Grading (ASAG) systems leverage Natural Language Processing (NLP) techniques to assess student responses and have been widely studied in various research communities~\cite{bonthu2021automated}. Recent advancements in Large Language Models (LLMs) have further facilitated the development of ASAG tools, making automatic grading more feasible and accessible.

Natural Language Processing (NLP), a subfield of artificial intelligence, focuses on training and deploying models that interpret and generate human language. A major breakthrough in NLP came in 2017 with the introduction of the Transformer architecture, which relies entirely on attention mechanisms to capture dependencies in input and output sequences~\cite{vaswani2023attention}. This architecture improved performance while reducing the computational resources needed for training and inference~\cite{vaswani2023attention}. Further developments led to the introduction of various pretrained LLMs, such as BERT\cite{devlin2019bert}, Llama\cite{touvron2023llama}, and OpenAI’s GPT family~\cite{brown2020gpt3}. These models have demonstrated strong performance across a variety of NLP tasks, often requiring minimal prompt engineering to adapt to new datasets~\cite{gilardi2023outperforms,kuzman2023beginning}.

More recently, LLMs have inspired the development of new ASAG tools, which generally follow one of two main approaches. The first approach utilizes numerical representations of student responses, transforming them into vector embeddings using LLMs and then applying techniques such as logistic regression, linear regression, or reference-based comparison to generate grades~\cite{fernandez2023automated,liu2019multiway,zhao2024autograding}. While these methods have demonstrated high grading accuracy across a range of educational levels, they require extensive training data and struggle to provide detailed, interpretable feedback~\cite{zhao2024autograding}. 
The second approach leverages generative models such as ChatGPT, where grades and feedback are produced through prompt engineering~\cite{impey2025using,kortemeyer2023performance,meyer2024combined}. This method enables more flexible feedback generation and adaptive instruction, making it appealing for formative assessment. These approaches, which did not include graded examples in their prompts, were not able to consistently match the performance of top performing custom-built models on standard ASAG datasets~\cite{bonthu2021automated,impey2025using,kortemeyer2023performance,meyer2024combined}.





\subsection{Few-Shot Training and Retrieval Augmented Generation}
\label{sec:related-rag}
One major issue related to the use of LLMs is hallucination, where models generate responses that are inaccurate or inconsistent with the intended grading criteria~\cite{huang2024hallucination}. Hallucination in ASAG can occur when the model lacks domain-specific knowledge, leading to incorrect grading and misleading feedback~\cite{grevisse2024medical}. Strategies such as few-shot learning and retrieval-augmented generation (RAG) have been explored to mitigate this issue. Few-shot learning provides models with example responses to guide grading, while RAG improves contextual grounding by retrieving relevant information from external sources~\cite{brown2020gpt3,lewis2021retrieval-augmented}. These techniques have shown promise in enhancing the model response accuracy and reducing hallucinations~\cite{lewis2021retrieval-augmented}. Few-shot learning has also been applied to ASAG. For example, Duong and Meng~\cite{duong2024automatic} incorporated graded examples and related course materials into prompts to provide additional context. However, their work does not compare different strategies for selecting these graded examples, leaving an open question about which selection methods are the most effective.

\subsection{Rubrics and Human Grading}
Another potentially critical factor in ASAG is the role of grading rubrics. Rubrics define evaluation criteria, quality benchmarks, and scoring strategies, making grading more transparent and interpretable~\cite{reddy2010rubric}. In formative assessment settings, rubrics can help students regulate their learning and improve performance~\cite{panadero2013formativerubric}. Instructors also use rubrics to maintain grading consistency across responses~\cite{reddy2010rubric}. 
In the ASAG contexts, Senanayake and Asanka~\cite{senanayake2024rubric-asag} integrated a grading rubric in the grading process, but they did not provide a complete analysis of the effect of the rubric. 

\vskip 8 pt

In this work, we address the existing knowledge gap regarding the impact of graded example selection and grading rubrics on ASAG performance. To build on these insights, we propose an ASAG pipeline that integrates LLM-based grading strategies. This paper presents the following key contributions:
\begin{enumerate}
\item Developing an ASAG pipeline leveraging existing Large Language Models.
\item Comparing the grading performance of three OpenAI generative models.
\item Examining the impact of integrating instructor-graded examples into the grading process, comparing three selection strategies: no examples, random selection, and RAG-based selection.
\item Analyzing the effect of incorporating grading rubrics on grading accuracy.
\end{enumerate}

\section{Grading method}
\subsection{Data}
\label{sec:method-data}
We utilize a number of public benchmark ASAG datasets created in previous works. 
Each dataset consists of a set of question prompts, a set of student submissions to the questions, and instructor-provided grades to these 
submissions that we can use as ground truth labels for the models. Some of the datasets contain reference solutions to these questions as well. 
The Texas dataset~\cite{mohler2011texas} includes questions from a Data Structure courses, with over 80 questions and $2{,}500$ graded submissions. 
SAF~\cite{filighera2022saf} covers a variety of college-level communication network topic, and has over 20 questions and $4{,}500$ graded submissions. 
The SciEntsBank dataset~\cite{nielsen2008scientsbank} focuses on 15 different science domains such as Life Science and Space Science, containing almost 200 questions and over $10{,}000$ submissions. 
Some other ASAG datasets such as Statistics~\cite{menini2019automated}, DigiKlausur~\cite{jeeveswaran2019digiklausurasag-dataset}, and Beetle~\cite{dzikovska2010beetle} are not used in this study due to the lack of question prompts, which form a fundamental part of our prompting method. 
These datasets worked for earlier models using trained classifiers as the question prompt is not needed for these models. 
We also utilize the mathematical induction proof dataset~\cite{poulsen2024student} for a comparison on the LLM's performance using rubrics. 
The Proof dataset contains over $3{,}000$ submissions on 4 induction questions. 
These proofs were graded using a 7-item rubric, with each item corresponding to a step of the induction proof. Each step is also related to a misconception in induction proofs that has been identified before~\cite{zhao2024autograding}. 

\begin{itemize}[leftmargin=.5in]
    \item[\textbf{R1}] Identifying the base case(s)
    \item[\textbf{R2}] Proving the base case(s)
    \item[\textbf{R3}] Stating the inductive hypothesis
    \item[\textbf{R4}] Setting the bound of the inductive hypothesis
    \item[\textbf{R5}] Stating the goal of the inductive step
    \item[\textbf{R6}] Breaking down the inductive step
    \item[\textbf{R7}] Applying the inductive hypothesis
\end{itemize}

A more detailed explanation on each rubric item can be found in \cite{poulsen2024student}. 
This rubric contributed to a high inter-rater reliability among human graders~\cite{poulsen2023measuring,poulsen2024disentangling,poulsen2023efficiency}. 

\subsection{Prompting}
\label{sec:prompting}


\begin{figure}
    \centering
    \includegraphics[width=\linewidth]{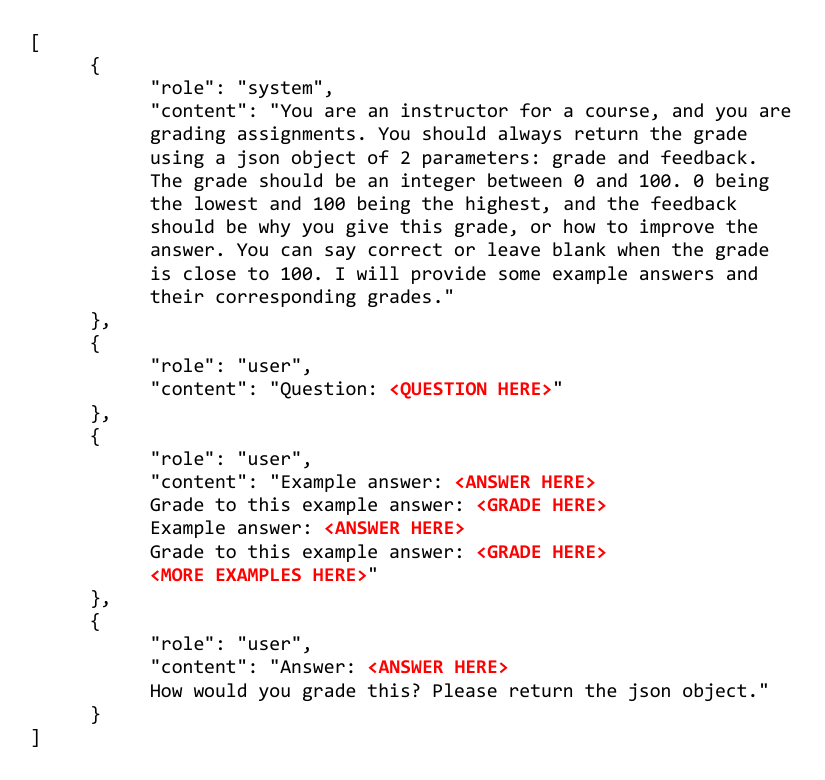}
    \caption{Our prompt for grading. The LLM is provided with generic grading instructions, the question that was asked to the students, some example graded solutions, and the solution that we are asking it to provide a grade for. Examples are selected with strategies explained in Section~\ref{sec:example-selection}.
}
    \label{fig:prompt-grading}
\end{figure} 

We use three LLM API endpoints throughout this study: \textit{gpt-4-turbo-2024-04-09}, \textit{gpt-4o-2024-08-06}, and \textit{o1-preview}. \textit{gpt-4-turbo-2024-04-09} (we will refer to it as GPT-4) is an older version of a high-intelligence GPT model. It can be run at a lower cost and can used to compare against other models. \textit{gpt-4o-2024-08-06} (GPT-4o) is the latest GPT-4o model at the time of the study. 
\textit{o1-preview} is the simplified and public version of \textit{o1}, the latest OpenAI model. It utilizes reinforcement learning and is able to perform a long chain of thought before responding~\cite{openai2024o1}. 

We design a prompt for LLMs aimed at achieving the best adaptability, as illustrated in Figure~\ref{fig:prompt-grading}. 
We divide the prompts into four components. The first component includes general grading instructions, 
offering the LLMs context on how to generate grading results and construct feedback. This part of the prompt is generic and remains unchanged for any course or dataset. 
The second component provides the question prompt from the datasets, enabling the LLM to generate its own correct answer for comparison. The third component consists of some graded examples from the dataset: a subset of human-graded responses selected by the system. 
The number of examples included can vary depending on instructor preferences, but using a larger number increases computational costs. 
For this study, we utilize 5 examples per prompt, selected using strategies detailed in Section~\ref{sec:example-selection}. 
Finally, the fourth component presents the student’s response to be graded.

Note that our grading prompt does not rely on sample problem solutions provided by the instructor.  In some classroom settings, a written reference solution may not always be available, and full-score graded examples can serve as effective substitutes for reference solutions. 

To enhance automation in the grading process, the LLM is instructed to return grading results in a structured JSON object. 
The JSON object contains 2 parameters: an integer ``grade'' between 0 and 100, and a string ``feedback'' with the feedback from the LLM. 
Since GPT-4 does not natively support structured output enforcement, we do not use structured output for GPT-4o and o1-preview either, to control for these conditions. As a result, there is no guaranteed way to ensure a 100\% JSON-compliant response. To address this, responses that deviate from the required JSON format are re-graded by the LLM. Fewer than 1\% of submissions produced invalid JSON objects, necessitating re-grading. 

\subsection{Selecting Graded Examples}
\label{sec:example-selection}
As each question in the datasets has multiple graded submissions, we can incorporate some submissions as examples for the LLM when grading another submission. We employ one of two selection strategies: \textit{Random} or \textit{RAG}. 
The \textit{Random} strategy randomly selects a fixed-size subset from the pool of human-graded submissions for the question and is easy to implement.
In contrast, the \textit{RAG} strategy leverages a retrieval-based approach introduced in Section~\ref{sec:related-rag}. It selects graded examples that are most similar to the current submission based on the Euclidean distance of their embeddings. 
While the \textit{Random} approach is computationally simpler, the \textit{RAG} method aims to maximize contextual relevance, potentially improving grading accuracy. 
As a baseline comparison, we add a \textit{None} strategy that does not include any graded examples. 
The \textit{None}, \textit{Random}, and \textit{RAG} strategies enable us to evaluate the effect of example selection on the performance of the grading pipeline.

\subsection{Grading with rubrics}
\label{sec:grading-rubrics}

\begin{figure}
    \centering
    \includegraphics[width=\linewidth]{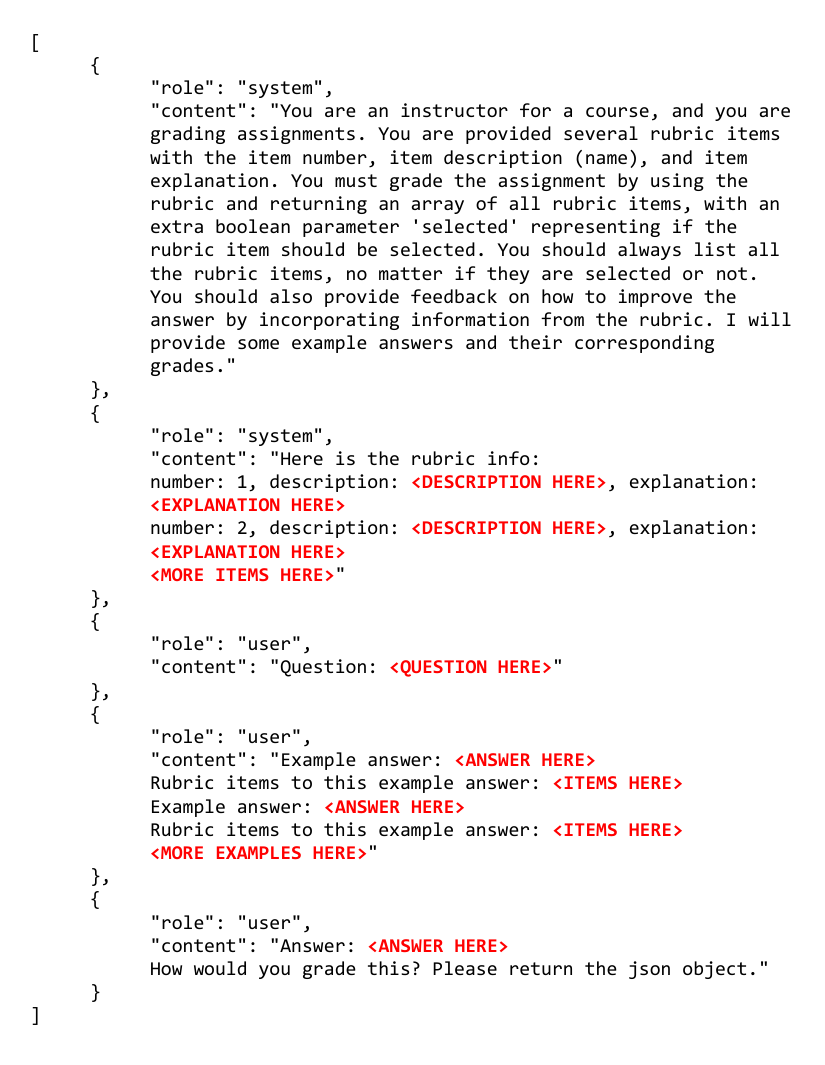}
    \caption{A prompt for grading using a rubric. Similar to Section \ref{sec:prompting}, the first component is the same across all courses and datasets. }
    \label{fig:prompt-grading-rubric}
\end{figure}

Our rubric-based prompt is adapted from the original grading prompt, 
enhanced with additional rubric-related information, as shown in Figure~\ref{fig:prompt-grading-rubric}. 
Each rubric item is assigned a unique number for system reference, 
a description or name presented to students, and a detailed explanation clarifying its meaning and criteria for correctness. 
By incorporating rubrics, the LLM transitions to making a series of binary decisions on each rubric item based on the explanation of the item, as opposed to assessing the overall quality of a student submission without a standardized framework. Scores can be computed based on the weights assigned to each rubric item by the instructor. The grading accuracy metric can be evaluated either by analyzing the accuracy of individual rubric items or by examining the overall calculated scores derived from the rubric.




To mitigate the risk of LLM hallucination, such as inventing new rubric items or failing to return them correctly, we utilize structured output enforcement in the GPT-4 model. The output includes a string labeled ``feedback'' and a ``rubric\_items'' object, which is a list of rubric items containing the name of each item and whether the item should be selected. This structured output ensures that the model consistently returns a list of results based on the provided rubric items, reducing the need for manual verification, particularly during large-scale deployment.

\section{Results}

\subsection{Comparison on models and effect of graded example selection method}
\label{sec:res-1}
\begin{figure}[ht]
    \centering
    \includegraphics[width=1\linewidth]{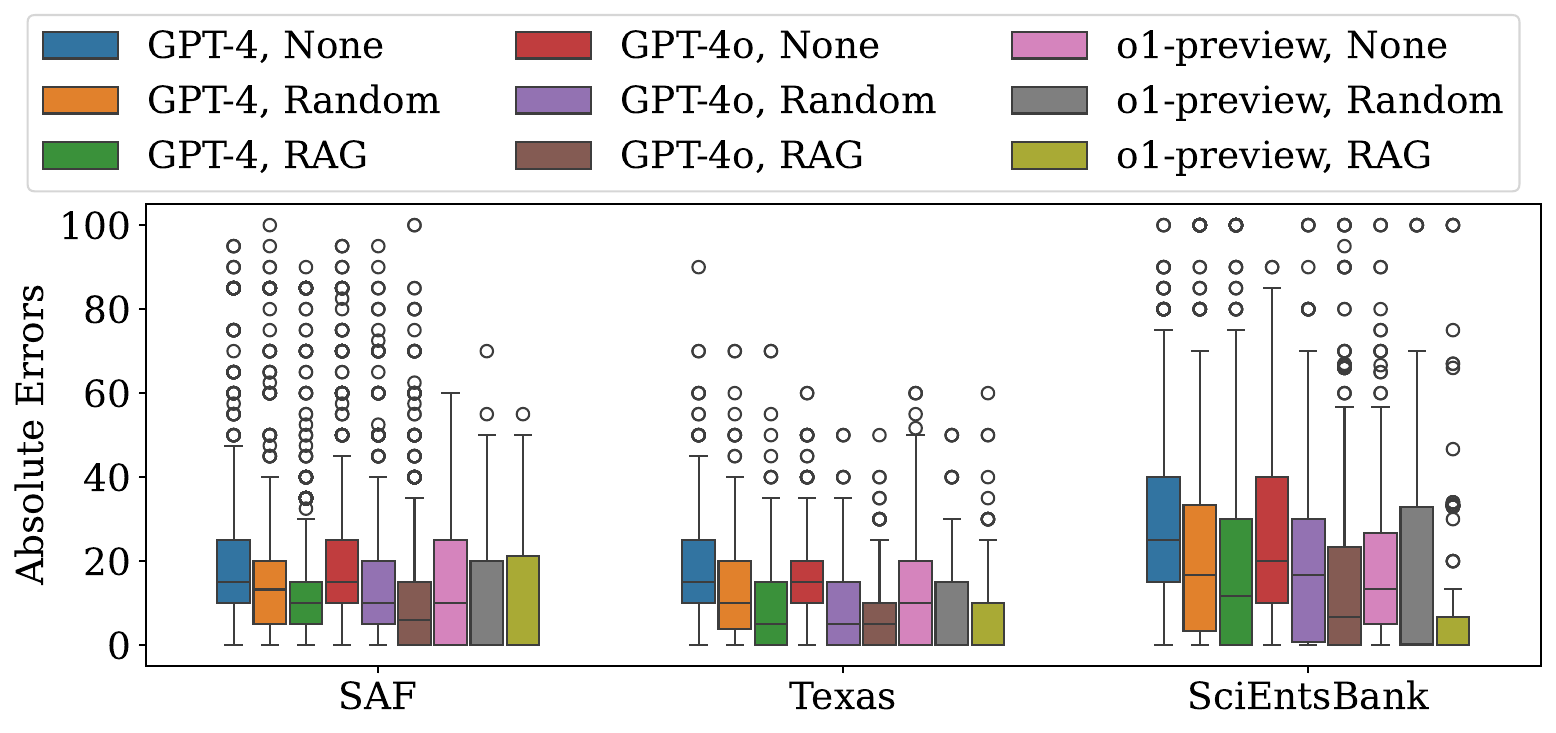}
    \caption{Boxplot showing the distribution of absolute errors for each dataset and grading method. The GPT-4 and GPT-4o models are run on the entire dataset, whereas the o1-preview model is run on 200 submissions at random. }
    \label{fig:model-errors}
\end{figure}

The Texas, SAF, and SciEntsBank datasets provide scores that are not based on rubrics and are on different grading scales. To ensure comparability on the accuracy metrics, we normalize these scores to percentages and then evaluate the grading performance using Root Mean Squared Error (RMSE) and Pearson's $r$ metrics. The results of these evaluations are presented in Tables~\ref{tab:accuracy-texas},~\ref{tab:accuracy-saf}, and~\ref{tab:accuracy-scientsbank} for Texas, SAF, and SciEntsBank respectively, 
and the distributions of absolute grading errors for each experiment group are shown in Figure~\ref{fig:model-errors}.

\begin{table}[h]
    \centering
        \begin{tabular}{ c ? C{1.6cm}  C{1.6cm}  C{1.6cm} ? C{1.6cm}  C{1.6cm}  C{1.6cm} } 
             & \multicolumn{3}{c?}{Pearson's $r$} & \multicolumn{3}{c}{RMSE} \\[2pt]

            & GPT-4 & GPT-4o & o1-preview & GPT-4 & GPT-4o & o1-preview \\[2pt] \specialrule{1pt}{0em}{0em} 

            None 
            & 0.56 & 0.64 & 0.55
            & 23.30 & 20.55 & 20.99 \\[2pt] \hline
            Random
            & 0.74 & 0.80 & 0.77
            & 17.82 & 14.94 & 15.31 \\[2pt] \hline
            RAG
            & 0.80 & 0.88 & 0.82
            & 14.79 & 11.73 & 13.26 \\[2pt]

        \end{tabular}
    \caption{Pearson's $r$ and RMSE metrics from the Texas dataset. We compare our results with two previous studies. In ~\cite{kumar2017earth}, Bi-LSTM resulted in a Pearson's $r$ of 0.64 and RMSE of 16.6. Another study using GPT-4o~\cite{meyer2024combined} without graded examples reported an RMSE of 22. Our pipeline including GPT-4o and graded examples improves performance metrics, without the need to train a model. }
    \label{tab:accuracy-texas}
\end{table}

\begin{table}[h]
    \centering
        \begin{tabular}{ c ? C{1.6cm}  C{1.6cm}  C{1.6cm} ? C{1.6cm}  C{1.6cm}  C{1.6cm} } 
             & \multicolumn{3}{c?}{Pearson's $r$} & \multicolumn{3}{c}{RMSE} \\[2pt]

            & GPT-4 & GPT-4o & o1-preview & GPT-4 & GPT-4o & o1-preview \\[2pt] \specialrule{1pt}{0em}{0em} 

            None 
            & 0.47 & 0.51 & 0.72
            & 27.46 & 26.80 & 20.61 \\[2pt] \hline
            Random
            & 0.63 & 0.72 & 0.81
            & 24.09 & 21.32 & 16.57 \\[2pt] \hline
            RAG
            & 0.68 & 0.80 & 0.78
            & 22.85 & 18.62 & 17.48 \\[2pt]

        \end{tabular}
    \caption{Pearson's $r$ and RMSE metrics from the SAF dataset. We compare our results with two previous studies. In ~\cite{filighera2022saf}, T5 resulted in an RMSE of 26.9. Another study using GPT-4o~\cite{meyer2024combined} without graded examples reported an RMSE of 24. Our pipeline including GPT-4o and graded examples improves performance metrics, without the need to train a model.}
    \label{tab:accuracy-saf}
\end{table}

\begin{table}[h]
    \centering
        \begin{tabular}{ c ? C{1.6cm}  C{1.6cm}  C{1.6cm} ? C{1.6cm}  C{1.6cm}  C{1.6cm} } 
             & \multicolumn{3}{c?}{Pearson's $r$} & \multicolumn{3}{c}{RMSE} \\[2pt]

            & GPT-4 & GPT-4o & o1-preview & GPT-4 & GPT-4o & o1-preview \\[2pt] \specialrule{1pt}{0em}{0em} 

            None 
            & 0.55 & 0.60 & 0.67
            & 33.58 & 31.79 & 29.52 \\[2pt] \hline
            Random
            & 0.67 & 0.70 & 0.72
            & 30.85 & 27.99 & 28.44 \\[2pt] \hline
            RAG
            & 0.71 & 0.77 & 0.81
            & 28.47 & 25.16 & 23.83 \\[2pt]

        \end{tabular}
    \caption{Pearson's $r$ and RMSE metrics from the SciEntsBank dataset. We compare our results with two previous studies. In ~\cite{filighera2022saf}, T5 resulted in an RMSE of 29.67. Another study using GPT-4o~\cite{meyer2024combined} without graded examples reported an RMSE of 31. Our pipeline including GPT-4o and graded examples improves performance metrics, without the need to train a model.}
    \label{tab:accuracy-scientsbank}
\end{table}

For both GPT-4 and GPT-4o, the time to grade each submission is around 1-2 seconds, and the cost is less than \$0.01. 
On the other hand, it takes o1-preview around 10 seconds to grade each submission, with the cost of around \$0.1 per grading task. 
Due to this difference, we run GPT-4 and 4o using the entire dataset, but only use a random sample of 200 submissions for o1-preview.
For all three datasets, the GPT-4o model without graded examples achieves comparable grading accuracy 
to the benchmark results presented in \cite{meyer2024combined} using GPT-4o. Notably, GPT-4o outperforms GPT-4 likely due to its enhanced knowledge base.
Based on the results in Figure~\ref{fig:model-errors}, over 50\% of the grading jobs performed by o1-preview achieve near-perfect results. 
This agrees with the promising performance of chain-of-thought methods observed in earlier studies~\cite{wei2023chain-of-thought}. 
However, there is also a larger variance in error for most experiment groups, agreeing with previous observations 
that reasoning paths can vary in quality~\cite{wang2024strategic,wei2023chain-of-thought}. 


To examine the effect of providing examples on absolute grading error, we combine all three datasets and conduct independent 
$t$-tests for each model. Within each model, incorporating few-shot learning through randomly selected graded examples significantly improves grading accuracy compared to not using graded examples ($t=10.84, p<0.001$ for GPT-4, $t=11.98, p<0.001$ for GPT-4o, $t=4.83, p<0.001$ for o1-preview). Furthermore, \textit{RAG}-based example selection is found to be even more effective in enhancing grading quality compared to \textit{Random} selection across all models ($t=5.21, p<0.001$ for GPT-4, $t=7.25, p<0.001$ for GPT-4o, $t=1.94, p=0.052$ for o1-preview), aligning with the effectiveness of RAG methods observed in other studies~\cite{lewis2021retrieval-augmented}. As a result, it is recommended that instructors grade some submissions initially so the model can use them as few-shot learning examples. However, \textit{RAG}-based example selection may require storing submission embeddings, which can lead to large storage requirements and increased computational costs when retrieving relevant examples. 

\subsection{The Effect of Using a Rubric}
We want to study the impact of using a validated ASAG rubric. 
We use the induction proof dataset for this comparison, which includes 4 different mathematical questions. The rubrics are described in Section~\ref{sec:method-data}. 
Since previous analysis in Section \ref{sec:res-1} suggests better accuracy with RAG-based example selection, we completed the grading tasks for the proof dataset by selecting 5 examples using retrieval. 
We use GPT-4o due to its lower cost compared to o1-preview and higher performance compared to GPT-4. For the grading with rubric, we use equal weights on the 7 rubric items and calculate the scores based on these weights. The RMSE and Pearson's $r$ were completed using these weighted scores, and are depicted in Table~\ref{tab:accuracy-rubric-vs-no-rubric-score}. 
The by-rubric-item accuracies are shown in Table~\ref{tab:accuracy-rubric-vs-no-rubric-item}. 


\begin{table}[h]
\centering
\begin{tabular}{ c  c  c  c } 

Question & Grading Method 
& Pearson's $r$ & RMSE \\ \toprule 

\multirow{3}{*}{P1} & GPT-3.5 Embeddings~\cite{zhao2024autograding} & 0.90 & 17.06 \\ \cmidrule(l){2-4}
& Human~\cite{zhao2024autograding} & 0.81 & 22.20 \\ \cmidrule(l){2-4}
& GPT-4o No rubric & 0.86 & 16.38 \\ \cmidrule(l){2-4}
& GPT-4o With rubric & 0.86 & 16.72 \\ \cmidrule(l){1-4}
 
\multirow{3}{*}{P2} & GPT-3.5 Embeddings~\cite{zhao2024autograding} & 0.86 & 22.32 \\ \cmidrule(l){2-4}
& Human~\cite{zhao2024autograding} & 0.94 & 13.58 \\ \cmidrule(l){2-4}
& GPT-4o No rubric & 0.91 & 16.67 \\ \cmidrule(l){2-4}
& GPT-4o With rubric & 0.92 & 15.91 \\ \cmidrule(l){1-4}

\multirow{3}{*}{P3} & GPT-3.5 Embeddings~\cite{zhao2024autograding} & 0.88 & 17.07 \\ \cmidrule(l){2-4}
& Human~\cite{zhao2024autograding} & 0.78 & 23.55 \\ \cmidrule(l){2-4}
& GPT-4o No rubric & 0.89 & 16.30 \\ \cmidrule(l){2-4}
& GPT-4o With rubric & 0.91 & 14.91 \\ \cmidrule(l){1-4}

\multirow{3}{*}{P4} & GPT-3.5 Embeddings~\cite{zhao2024autograding} & 0.83 & 17.86 \\ \cmidrule(l){2-4}
& Human~\cite{zhao2024autograding} & 0.71 & 25.56 \\ \cmidrule(l){2-4}
& GPT-4o No rubric & 0.86 & 17.60 \\ \cmidrule(l){2-4}
& GPT-4o With rubric & 0.92 & 13.67 \\ \cmidrule(l){1-4}

\end{tabular}
\caption{Grading accuracy for 4 mathematical induction questions}
\label{tab:accuracy-rubric-vs-no-rubric-score}
\end{table}

Including rubric information results in a significantly higher grading accuracy for two of the questions questions, with $t=2.20, p=0.03$ for question 3 (P3), and $t=4.64, p<0.001$ for question 4 (P4). We observe negligible influence for the other two questions, with $t=1.34, p=0.18$ for question 1 (P1), and $t=1.31, p=0.19$ for question 2 (P2). Overall, providing a rubric generally aids the LLM in generating more accurate grades. 
Previous studies on ASAG and human grading~\cite{fowler2021eipe,zhao2024autograding} have shown that achieving 100\% agreement among graders is nearly impossible due to the open-ended nature of the questions and the ambiguity of natural language. Incorporating a rubric can help LLMs achieve greater consistency and alignment with human grading standards.
Interestingly, the by-rubric-item accuracy of GPT-4o is already comparable to that of GPT-3.5-based task-specific grading models as well as human graders recruited in earlier studies, highlighting the remarkable capability of GPT-4o.

\begin{table}[h]
\centering
\begin{tabular}{ c  c  c } 

Question & Grading Method & Item accuracy \\ \toprule 

\multirow{3}{*}{P1} & GPT-3.5 Embeddings~\cite{zhao2024autograding} & 87.9\% \\ \cmidrule(l){2-3}
& Human~\cite{zhao2024autograding} & 84.1\% \\ \cmidrule(l){2-3}
& GPT-4o With rubric & 86.4\%\\ \cmidrule(l){1-3}
 
\multirow{3}{*}{P2} & GPT-3.5 Embeddings~\cite{zhao2024autograding} & 88.3\% \\ \cmidrule(l){2-3}
& Human~\cite{zhao2024autograding} & 91.3\% \\ \cmidrule(l){2-3}
& GPT-4o With rubric & 87.7\%\\ \cmidrule(l){1-3}

\multirow{3}{*}{P3} & GPT-3.5 Embeddings~\cite{zhao2024autograding} & 89.2\% \\ \cmidrule(l){2-3}
& Human~\cite{zhao2024autograding} & 85.1\% \\ \cmidrule(l){2-3}
& GPT-4o With rubric & 89.6\%\\ \cmidrule(l){1-3}

\multirow{3}{*}{P4} & GPT-3.5 Embeddings~\cite{zhao2024autograding} & 85.4\% \\ \cmidrule(l){2-3}
& Human~\cite{zhao2024autograding} & 85.8\% \\ \cmidrule(l){2-3}
& GPT-4o With rubric & 89.9\% \\ \cmidrule(l){1-3}

\end{tabular}
\caption{Rubric item accuracy for 4 mathematical induction questions}
\label{tab:accuracy-rubric-vs-no-rubric-item}
\end{table}

\section{Conclusions and Future Work}
In this work, we develop an Automatic Short Answer Grading pipeline by prompting a Large Language Model to generate a score and evaluate its grading performance. 
Combining the use of GPT-4o, RAG-based example selection, and a grading rubric, the grading pipeline developed in this work achieves state of the art performance on all datasets with the exception of a few complex mathematical proof problems.
The system also has similar accuracy human graders~\cite{zhao2024autograding,fowler2021eipe}.

Among the OpenAI's GPT-4, GPT-4o, and o1-preview models, GPT-4o currently has the best performance on the grading tasks in terms of accuracy and cost. 
Although o1-preview produces a higher percentage of accurate grades, its high variance in error would make it unfit for real classroom settings. 
The cost for o1-preview is too high for scaled use as well. 
Further-developed chain-of-thought strategies could potentially make o1-preview a more suitable model for classroom use. 
For the grading tasks, providing graded examples in the prompt can effectively increase the grading accuracy, 
and choosing examples using RAG-based methods has a larger impact than randomly selecting examples. 
Providing a rubric in the grading pipeline can also increase grading accuracy by offering interpretability in the grading process.

Given the high accuracy, our grading pipeline can be used in real classroom settings to help decrease the workload on human graders. An appeal system could be used to mitigate the effects of grading mistakes, similar to how appeal systems are used to mitigate error in human TA grading.
A future direction is to integrate the grading pipeline into an open question-answering platform with an interactive interface that allows instructors to actively assess the model's accuracy, determine its suitability for their use case, and provide additional graded examples as needed to refine performance. 
It can also be helpful for the instructors to customize the the grading jobs, such as by providing their own graded examples instead of only student submissions to address more specific misconceptions in the questions. 
The impact of such customizations can also be studied. 
Another direction for future work will be to address a limitation in our study: assessing the quality of the feedback generated by the grading pipeline.
This can be achieved by analyzing the relationship between the generated grades or rubric items and the corresponding feedback, 
as well as studying the impact of this feedback on student learning outcomes.

%
%
%
\bibliographystyle{splncs04}
\bibliography{references}

\begin{thebibliography}{10}
\providecommand{\url}[1]{\texttt{#1}}
\providecommand{\urlprefix}{URL }
\providecommand{\doi}[1]{https://doi.org/#1}

\bibitem{bonthu2021automated}
Bonthu, S., Sripada, R.S., Prasad, M.: Automated {Short} {Answer} {Grading} {Using} {Deep} {Learning}: {A} {Survey}. pp. 61--78 (Aug 2021). \doi{10.1007/978-3-030-84060-0_5}

\bibitem{brown2020gpt3}
Brown, T.B., Mann, B., Ryder, N., Subbiah, M., Kaplan, J., Dhariwal, P., Neelakantan, A., Shyam, P., Sastry, G., Askell, A., Agarwal, S., Herbert-Voss, A., Krueger, G., Henighan, T., Child, R., Ramesh, A., Ziegler, D.M., Wu, J., Winter, C., Hesse, C., Chen, M., Sigler, E., Litwin, M., Gray, S., Chess, B., Clark, J., Berner, C., McCandlish, S., Radford, A., Sutskever, I., Amodei, D.: Language {Models} are {Few}-{Shot} {Learners} (Jul 2020), \url{http://arxiv.org/abs/2005.14165}, arXiv:2005.14165 [cs]

\bibitem{devlin2019bert}
Devlin, J., Chang, M.W., Lee, K., Toutanova, K.: {BERT}: {Pre}-training of {Deep} {Bidirectional} {Transformers} for {Language} {Understanding} (May 2019), \url{http://arxiv.org/abs/1810.04805}, arXiv:1810.04805 [cs]

\bibitem{duong2024automatic}
Duong, T.N.B., Meng, C.Y.: Automatic {Grading} of {Short} {Answers} {Using} {Large} {Language} {Models} in {Software} {Engineering} {Courses}. In: 2024 {IEEE} {Global} {Engineering} {Education} {Conference} ({EDUCON}). pp. 1--10 (May 2024). \doi{10.1109/EDUCON60312.2024.10578839}, \url{https://ieeexplore.ieee.org/document/10578839}, iSSN: 2165-9567

\bibitem{dzikovska2010beetle}
Dzikovska, M.O., Moore, J.D., Steinhauser, N., Campbell, G., Farrow, E., Callaway, C.B.: Beetle {II}: {A} {System} for {Tutoring} and {Computational} {Linguistics} {Experimentation}. In: Kübler, S. (ed.) Proceedings of the {ACL} 2010 {System} {Demonstrations}. pp. 13--18. Association for Computational Linguistics, Uppsala, Sweden (Jul 2010), \url{https://aclanthology.org/P10-4003/}

\bibitem{fernandez2023automated}
Fernandez, N., Ghosh, A., Liu, N., Wang, Z., Choffin, B., Baraniuk, R., Lan, A.: Automated {Scoring} for {Reading} {Comprehension} via {In}-context {BERT} {Tuning} (Jun 2023). \doi{10.48550/arXiv.2205.09864}, \url{http://arxiv.org/abs/2205.09864}, arXiv:2205.09864 [cs]

\bibitem{filighera2022saf}
Filighera, A., Parihar, S., Steuer, T., Meuser, T., Ochs, S.: Your {Answer} is {Incorrect}... {Would} you like to know why? {Introducing} a {Bilingual} {Short} {Answer} {Feedback} {Dataset}. In: Muresan, S., Nakov, P., Villavicencio, A. (eds.) Proceedings of the 60th {Annual} {Meeting} of the {Association} for {Computational} {Linguistics} ({Volume} 1: {Long} {Papers}). pp. 8577--8591. Association for Computational Linguistics, Dublin, Ireland (May 2022). \doi{10.18653/v1/2022.acl-long.587}, \url{https://aclanthology.org/2022.acl-long.587/}

\bibitem{fowler2021eipe}
Fowler, M., Chen, B., Azad, S., West, M., Zilles, C.: Autograding "{Explain} in {Plain} {English}" questions using {NLP}. In: Proceedings of the 52nd {ACM} {Technical} {Symposium} on {Computer} {Science} {Education}. pp. 1163--1169. ACM, Virtual Event USA (Mar 2021). \doi{10.1145/3408877.3432539}, \url{https://dl.acm.org/doi/10.1145/3408877.3432539}

\bibitem{gilardi2023outperforms}
Gilardi, F., Alizadeh, M., Kubli, M.: Chatgpt outperforms crowd workers for text-annotation tasks. Proceedings of the National Academy of Sciences  \textbf{120}(30) (Jul 2023). \doi{10.1073/pnas.2305016120}, \url{http://dx.doi.org/10.1073/pnas.2305016120}

\bibitem{grevisse2024medical}
Grévisse, C.: {LLM}-based automatic short answer grading in undergraduate medical education. BMC Medical Education  \textbf{24}(1), ~1060 (Sep 2024). \doi{10.1186/s12909-024-06026-5}, \url{https://doi.org/10.1186/s12909-024-06026-5}

\bibitem{ashish2024tradeoff}
Gurung, A., Vanacore, K., Mcreynolds, A.A., Ostrow, K.S., Worden, E., Sales, A.C., Heffernan, N.T.: Multiple choice vs. fill-in problems: The trade-off between scalability and learning. In: Proceedings of the 14th Learning Analytics and Knowledge Conference. p. 507–517. LAK '24, Association for Computing Machinery, New York, NY, USA (2024). \doi{10.1145/3636555.3636908}, \url{https://doi.org/10.1145/3636555.3636908}

\bibitem{henkel2024can}
Henkel, O., Hills, L., Boxer, A., Roberts, B., Levonian, Z.: Can {Large} {Language} {Models} {Make} the {Grade}? {An} {Empirical} {Study} {Evaluating} {LLMs} {Ability} {To} {Mark} {Short} {Answer} {Questions} in {K}-12 {Education}. In: Proceedings of the {Eleventh} {ACM} {Conference} on {Learning} @ {Scale}. pp. 300--304. L@{S} '24, Association for Computing Machinery, New York, NY, USA (Jul 2024). \doi{10.1145/3657604.3664693}, \url{https://dl.acm.org/doi/10.1145/3657604.3664693}

\bibitem{huang2024hallucination}
Huang, L., Yu, W., Ma, W., Zhong, W., Feng, Z., Wang, H., Chen, Q., Peng, W., Feng, X., Qin, B., Liu, T.: A survey on hallucination in large language models: Principles, taxonomy, challenges, and open questions. ACM Transactions on Information Systems  (Nov 2024). \doi{10.1145/3703155}, \url{http://dx.doi.org/10.1145/3703155}

\bibitem{impey2025using}
Impey, C., Wenger, M., Garuda, N., Golchin, S., Stamer, S.: Using {Large} {Language} {Models} for {Automated} {Grading} of {Student} {Writing} about {Science}. Int J Artif Intell Educ  (Jan 2025). \doi{10.1007/s40593-024-00453-7}, \url{http://arxiv.org/abs/2412.18719}, arXiv:2412.18719 [cs]

\bibitem{jeeveswaran2019digiklausurasag-dataset}
Jeeveswaran, K.: {DigiKlausur}/{ASAG}-{Dataset} (2019), \url{https://github.com/DigiKlausur/ASAG-Dataset}, original-date: 2019-02-26T10:22:22Z

\bibitem{kortemeyer2023performance}
Kortemeyer, G.: Performance of the {Pre}-{Trained} {Large} {Language} {Model} {GPT}-4 on {Automated} {Short} {Answer} {Grading} (Sep 2023). \doi{10.48550/arXiv.2309.09338}, \url{http://arxiv.org/abs/2309.09338}, arXiv:2309.09338 [cs]

\bibitem{kumar2017earth}
Kumar, S., Chakrabarti, S., Roy, S.: Earth {Mover}'s {Distance} {Pooling} over {Siamese} {LSTMs} for {Automatic} {Short} {Answer} {Grading} pp. 2046--2052 (2017), \url{https://www.ijcai.org/proceedings/2017/284}

\bibitem{kuzman2023beginning}
Kuzman, T., Mozetič, I., Ljubešić, N.: Chatgpt: Beginning of an end of manual linguistic data annotation? use case of automatic genre identification (2023), \url{https://arxiv.org/abs/2303.03953}

\bibitem{lewis2021retrieval-augmented}
Lewis, P., Perez, E., Piktus, A., Petroni, F., Karpukhin, V., Goyal, N., Küttler, H., Lewis, M., Yih, W.t., Rocktäschel, T., Riedel, S., Kiela, D.: Retrieval-{Augmented} {Generation} for {Knowledge}-{Intensive} {NLP} {Tasks} (Apr 2021). \doi{10.48550/arXiv.2005.11401}, \url{http://arxiv.org/abs/2005.11401}, arXiv:2005.11401 [cs]

\bibitem{liu2019multiway}
Liu, T., Ding, W., Wang, Z., Tang, J., Huang, G.Y., Liu, Z.: Automatic short answer grading via multiway attention networks (2019), \url{https://arxiv.org/abs/1909.10166}

\bibitem{menini2019automated}
Menini, S., Tonelli, S., Gasperis, G.D., Vittorini, P.: Automated {Short} {Answer} {Grading}: {A} {Simple} {Solution} for a {Difficult} {Task}. In: Italian Conference on Computational Linguistics (2019), \url{https://api.semanticscholar.org/CorpusID:204922864}

\bibitem{meyer2024combined}
Meyer, G., Breuer, P., Fürst, J.: {ASAG2024}: {A} {Combined} {Benchmark} for {Short} {Answer} {Grading}. In: Proceedings of the 2024 on {ACM} {Virtual} {Global} {Computing} {Education} {Conference} {V}. 2. pp. 322--323. {SIGCSE} {Virtual} 2024, Association for Computing Machinery, New York, NY, USA (Dec 2024). \doi{10.1145/3649409.3691083}, \url{https://dl.acm.org/doi/10.1145/3649409.3691083}

\bibitem{mohler2011texas}
Mohler, M., Bunescu, R., Mihalcea, R.: Learning to {Grade} {Short} {Answer} {Questions} using {Semantic} {Similarity} {Measures} and {Dependency} {Graph} {Alignments}. In: Lin, D., Matsumoto, Y., Mihalcea, R. (eds.) Proceedings of the 49th {Annual} {Meeting} of the {Association} for {Computational} {Linguistics}: {Human} {Language} {Technologies}. pp. 752--762. Association for Computational Linguistics, Portland, Oregon, USA (Jun 2011), \url{https://aclanthology.org/P11-1076/}

\bibitem{nielsen2008scientsbank}
Nielsen, R.D., Ward, W., Martin, J., Palmer, M.: Annotating {Students}' {Understanding} of {Science} {Concepts}. In: Calzolari, N., Choukri, K., Maegaard, B., Mariani, J., Odijk, J., Piperidis, S., Tapias, D. (eds.) Proceedings of the {Sixth} {International} {Conference} on {Language} {Resources} and {Evaluation} ({LREC}`08). European Language Resources Association (ELRA), Marrakech, Morocco (May 2008), \url{https://aclanthology.org/L08-1166/}

\bibitem{openai2024o1}
OpenAI: Learning to reason with {LLMs}, \url{https://openai.com/index/learning-to-reason-with-llms/}

\bibitem{panadero2013formativerubric}
Panadero, E., Jonsson, A.: The use of scoring rubrics for formative assessment purposes revisited: {A} review. Educational Research Review  \textbf{9},  129--144 (Jun 2013). \doi{10.1016/j.edurev.2013.01.002}, \url{https://www.sciencedirect.com/science/article/pii/S1747938X13000109}

\bibitem{pearson2005assessment}
Pearson, P.D., Hamm, D.N.: The {Assessment} of {Reading} {Comprehension}: {A} {Review} of {Practices}-{Past}, {Present}, and {Future}. In: Children's reading comprehension and assessment, pp. 13--69. Center for improvement of early reading achievement ({CIERA}), Lawrence Erlbaum Associates Publishers, Mahwah, NJ, US (2005)

\bibitem{poulsen2024student}
Poulsen, S.: {Student Proof by Induction Data Set} (2024). \doi{10.7910/DVN/OTRLXF}, \url{https://doi.org/10.7910/DVN/OTRLXF}

\bibitem{poulsen2023measuring}
Poulsen, S., Chen, H., Gertner, Y., Cosman, B., West, M., Herman, G.L.: Measuring the impact of distractors on student learning gains while using proof blocks (2023)

\bibitem{poulsen2024disentangling}
Poulsen, S., Gertner, Y., Chen, H., Cosman, B., West, M., Herman, G.L.: Disentangling the learning gains from reading a book chapter and completing proof blocks problems. In: Proceedings of the 55th ACM Technical Symposium on Computer Science Education V. 1 (2024)

\bibitem{poulsen2023efficiency}
Poulsen, S., Gertner, Y., Cosman, B., West, M., Herman, G.L.: Efficiency of learning from proof blocks versus writing proofs. In: Proceedings of the 54th ACM Technical Symposium on Computer Science Education V. 1. pp. 472--478 (2023)

\bibitem{reddy2010rubric}
Reddy, Y.M., Andrade, H.: A review of rubric use in higher education. Assessment \& Evaluation in Higher Education  \textbf{35}(4),  435--448 (2010). \doi{10.1080/02602930902862859}, \url{https://doi.org/10.1080/02602930902862859}

\bibitem{scott2006evaluating}
Scott, M., Stelzer, T., Gladding, G.: Evaluating multiple-choice exams in large introductory physics courses. Phys. Rev. ST Phys. Educ. Res.  \textbf{2}(2),  020102 (Jul 2006). \doi{10.1103/PhysRevSTPER.2.020102}, \url{https://link.aps.org/doi/10.1103/PhysRevSTPER.2.020102}, publisher: American Physical Society

\bibitem{senanayake2024rubric-asag}
Senanayake, C., Asanka, D.: Rubric {Based} {Automated} {Short} {Answer} {Scoring} using {Large} {Language} {Models} ({LLMs}). In: 2024 {International} {Research} {Conference} on {Smart} {Computing} and {Systems} {Engineering} ({SCSE}). vol.~7, pp.~1--6 (Apr 2024). \doi{10.1109/SCSE61872.2024.10550624}, \url{https://ieeexplore.ieee.org/abstract/document/10550624}, iSSN: 2613-8662

\bibitem{shute2008feedback}
Shute, V.J.: Focus on formative feedback. Review of Educational Research  \textbf{78}(1),  153--189 (2008). \doi{10.3102/0034654307313795}, \url{https://doi.org/10.3102/0034654307313795}

\bibitem{touvron2023llama}
Touvron, H., Martin, L., Stone, K., Albert, P., Almahairi, A., Babaei, Y., Bashlykov, N., Batra, S., Bhargava, P., Bhosale, S., Bikel, D., Blecher, L., Ferrer, C.C., Chen, M., Cucurull, G., Esiobu, D., Fernandes, J., Fu, J., Fu, W., Fuller, B., Gao, C., Goswami, V., Goyal, N., Hartshorn, A., Hosseini, S., Hou, R., Inan, H., Kardas, M., Kerkez, V., Khabsa, M., Kloumann, I., Korenev, A., Koura, P.S., Lachaux, M.A., Lavril, T., Lee, J., Liskovich, D., Lu, Y., Mao, Y., Martinet, X., Mihaylov, T., Mishra, P., Molybog, I., Nie, Y., Poulton, A., Reizenstein, J., Rungta, R., Saladi, K., Schelten, A., Silva, R., Smith, E.M., Subramanian, R., Tan, X.E., Tang, B., Taylor, R., Williams, A., Kuan, J.X., Xu, P., Yan, Z., Zarov, I., Zhang, Y., Fan, A., Kambadur, M., Narang, S., Rodriguez, A., Stojnic, R., Edunov, S., Scialom, T.: Llama 2: {Open} {Foundation} and {Fine}-{Tuned} {Chat} {Models} (Jul 2023), \url{http://arxiv.org/abs/2307.09288}, arXiv:2307.09288 [cs]

\bibitem{vaswani2023attention}
Vaswani, A., Shazeer, N., Parmar, N., Uszkoreit, J., Jones, L., Gomez, A.N., Kaiser, L., Polosukhin, I.: Attention {Is} {All} {You} {Need} (Aug 2023). \doi{10.48550/arXiv.1706.03762}, \url{http://arxiv.org/abs/1706.03762}, arXiv:1706.03762 [cs]

\bibitem{wang2024strategic}
Wang, Y., Zhao, S., Wang, Z., Huang, H., Fan, M., Zhang, Y., Wang, Z., Wang, H., Liu, T.: Strategic {Chain}-of-{Thought}: {Guiding} {Accurate} {Reasoning} in {LLMs} through {Strategy} {Elicitation} (Sep 2024). \doi{10.48550/arXiv.2409.03271}, \url{http://arxiv.org/abs/2409.03271}, arXiv:2409.03271 [cs]

\bibitem{wei2023chain-of-thought}
Wei, J., Wang, X., Schuurmans, D., Bosma, M., Ichter, B., Xia, F., Chi, E., Le, Q., Zhou, D.: Chain-of-{Thought} {Prompting} {Elicits} {Reasoning} in {Large} {Language} {Models} (Jan 2023). \doi{10.48550/arXiv.2201.11903}, \url{http://arxiv.org/abs/2201.11903}, arXiv:2201.11903 [cs]

\bibitem{zhao2024autograding}
Zhao, C., Silva, M., Poulsen, S.: Autograding mathematical induction proofs with natural language processing (2024), \url{https://arxiv.org/abs/2406.10268}

\end{thebibliography}
\end{document}